\documentclass[letterpaper, 10 pt, conference]{ieeeconf}

\IEEEoverridecommandlockouts
\usepackage{amsmath}
\usepackage{amssymb}
\usepackage{graphicx}
\usepackage{bm}

\makeatletter
\g@addto@macro\normalsize{%
  \setlength{\abovedisplayskip}{3pt plus 1pt minus 1pt}%
  \setlength{\belowdisplayskip}{3pt plus 1pt minus 1pt}%
  \setlength{\abovedisplayshortskip}{1pt plus 1pt}%
  \setlength{\belowdisplayshortskip}{2pt plus 1pt}}
\makeatother
\usepackage{booktabs}
\usepackage{tabularx}
\usepackage{xcolor}
\usepackage{cite}

\newcommand{\qref}{\bm{q}^{\star}}
\newcommand{\Rset}{\mathcal{R}}

\title{\LARGE \bf
Burden, Bounty, or Bliss? Muscle-Aware Retargeting and\\
Null-Space Exploration for Whole-Body Musculoskeletal Control
}

\author{Ruiyang Kong$^{1}$ and Takeshi Oishi$^{1}$%
\thanks{$^{1}$The University of Tokyo, Tokyo, Japan. This work has been submitted to the IEEE for possible publication. Copyright may be transferred without notice, after which this version may no longer be accessible.}%
}

\begin{document}

\maketitle
\thispagestyle{empty}
\pagestyle{empty}

\begin{abstract}
Learning whole-body musculoskeletal control from human motion is important for developing human-like robots, assistive devices, and rehabilitation systems.
This task presents two main challenges, namely reference trajectories that are kinematically plausible but difficult to reproduce through muscle-driven dynamics, and a highly redundant 700-dimensional action space in which unstructured exploration can easily disrupt the target motion.
In this paper, we propose a motion-to-policy pipeline that exploits muscle redundancy in two complementary stages.
First, \emph{muscle-aware retargeting} (MAR) performs bounded adaptation of kinematic trajectories subject to hard semantic, anatomical, and contact constraints.
We motivate the constraint bounds by analyzing how reference-trajectory artifacts affect muscle-driven tracking.
Second, \emph{null-space exploration} (NSE) estimates a one-control-frame mapping from muscle excitations to generalized forces and injects bounded, anatomically structured perturbations along its null space into the actions of a value-guided flow policy.
On a 24-clip motion-capture benchmark, MAR and NSE together solve every reference, reach the full horizon with half the samples of the baseline pipeline, and lower joint-center error by $44\%$ without late collapse.
To the best of our knowledge, this is the first end-to-end pipeline that transforms monocular video into a trained whole-body imitation policy controlling 700 muscles.
\end{abstract}

\section{Introduction}
Driving whole-body musculoskeletal models from human motion benefits biomechanics, embodied intelligence, and rehabilitation, but it is much harder than controlling torque-actuated humanoids.
MS-Human-700~\cite{zuo2024selfmodel}, for example, actuates $90$ body segments through $700$ muscle--tendon units that can only pull, and first-order activation dynamics low-pass filter every command, which complicates both reference generation and policy learning.

The first challenge is the \emph{retargeting gap}.
Inverse kinematics can align keypoints, yet small errors in foot height, contact timing, or joint orientation are amplified under muscle actuation.
Because activation acts as a low-pass filter, tracking high-frequency motion can require over ten times the control effort, and persistent reward biases distort the discounted return.

The second challenge is the \emph{exploration gap}.
In the 700-dimensional muscle space, isotropic noise directly perturbs generalized forces, policy-gradient variance grows with the action dimension, and early exploration falls repeatedly.
QFlex~\cite{wei2026qflex} scales value-guided flow exploration to this dimensionality but does not distinguish mechanically consequential action directions from near-null ones.

Our key insight is that \emph{muscle redundancy is not only a burden but also a bounty}.
Human movement variability concentrates along directions that barely affect the task, so diverse activation patterns can be explored without disrupting the motion~\cite{scholz1999ucm,todorov2002ofc,wu2014variability,dhawale2017variability,latash2012abundance}.
During retargeting, the search can target trajectories that preserve the source semantics yet are easier to realize with muscles.
During learning, exploration can be directed into the null space of excitation directions that barely affect short-term body motion.

\begin{figure*}[t]
  \centering
  \includegraphics[width=0.85\linewidth]{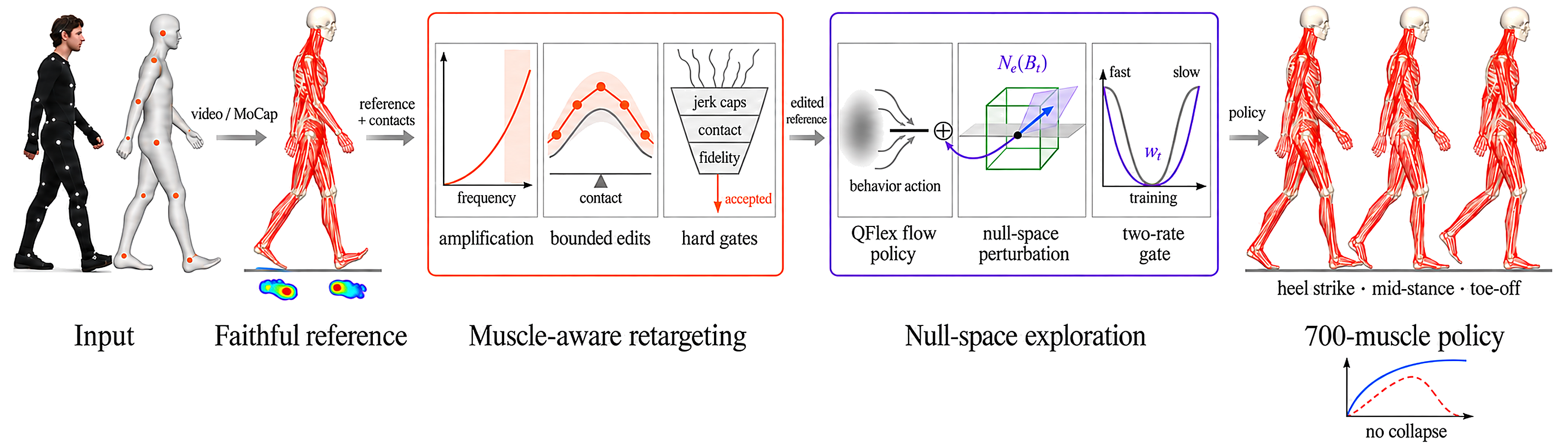}
  \caption{Pipeline from MoCap or video (GVHMR) through faithful IK and grounding, \textbf{MAR} with bounded edits under muscle-motivated gates, and QFlex imitation with \textbf{NSE} on the behavior actions to a 700-muscle policy.  Redundancy is used twice, to pick a realizable reference and to explore without disturbing the task.}
  \label{fig:pipeline}
\end{figure*}

We instantiate this insight in a motion-to-policy pipeline (Fig.~\ref{fig:pipeline}).
Motion from capture or video (GVHMR~\cite{shen2024gvhmr}) becomes a faithful, anatomically grounded reference.
Muscle-aware retargeting (MAR) then makes bounded low-dimensional edits under hard anatomical and contact constraints with muscle-informed budgets, and QFlex imitates the result while null-space exploration (NSE) perturbs its behavior actions along the null space of a one-control-frame excitation-to-force mapping.

This paper makes three contributions:
\begin{enumerate}
\item \textbf{Muscle-aware retargeting.}
A kinematic retargeting method that makes faithful motion easier to realize with muscles while preserving joint-center motion and contact structure, with edits bounded by hard continuity, frequency, and contact constraints informed by activation dynamics.
\item \textbf{Null-space exploration.}
An anatomy-structured exploration method that estimates a short-horizon excitation-to-force mapping and perturbs the QFlex behavior distribution along its approximate null space, enabling diverse exploration with limited task-space disturbance.
\item \textbf{An end-to-end video-to-policy pipeline.}
A naive chain of video pose estimation and QFlex fails because reference errors impose excessive control costs, whereas MAR and NSE bridge this gap (\mbox{Sec.~\ref{sec:videopolicy}}).
\end{enumerate}

\section{Related Work}

\subsection{Video-Based Human Motion Recovery}
Monocular human mesh recovery has advanced from parametric body-model regression~\cite{loper2015smpl,pavlakos2019smplx,goel2023humans4d} to global trajectory recovery~\cite{wang2024tram} and world-grounded estimation~\cite{shin2024wham,shen2024gvhmr}, and we use GVHMR~\cite{shen2024gvhmr} as our front end.
Physics-aware estimators impose plausibility by tracking the estimate with a simulated torque-actuated character~\cite{shimada2020physcap,yuan2021simpoe}, yet world-grounded estimates still contain foot skating, scale drift, and contact-timing errors, which MAR treats as a retargeting problem rather than a control problem.
Monocular video has driven torque-humanoid policies~\cite{allshire2025videomimic}, markerless capture supports muscle-driven clinical analysis~\cite{cotton2025kintwin}, and video biomechanics estimates muscle activations without learning a policy~\cite{uhlrich2023opencap}.

\subsection{Physics-Aware Motion Retargeting}
Conventional retargeting adapts motion across skeletons~\cite{gleicher1998retarget} and feeds humanoid teleoperation and imitation~\cite{he2024h2o,fu2024humanplus}.
Physics-based character animation instead absorbs reference artifacts during simulation-in-the-loop training~\cite{peng2018deepmimic,peng2021amp,luo2023phc,tessler2024maskedmimic}, relying on early termination and reference-state initialization~\cite{peng2018deepmimic} or discarding infeasible clips~\cite{luo2023phc} rather than repairing the reference.
We separate \emph{faithfulness}, \emph{physical validity}, and \emph{control feasibility}, the last meaning realizability under bounded and rate-limited activations, which MAR improves through muscle-informed bounds and hard constraints (Appendix~\ref{sec:amplification}) with minimal edits.

\subsection{High-Dimensional Musculoskeletal Control}
Platforms and systems include OpenSim~\cite{delp2007opensim}, MyoSuite~\cite{caggiano2022myosuite}, evolutionary~\cite{geijtenbeek2013muscle} and hierarchical~\cite{lee2019mass} muscle coordination, MS-Human-700~\cite{zuo2024selfmodel}, and tendon-driven musculoskeletal humanoids~\cite{kawaharazuka2026muscles}.
Recent work scales muscle-driven motion imitation~\cite{simos2026kinesis,li2026musclemimic,wei2026msemulator,wang2026myochallenge}, but its retargeting remains kinematic, its exploration generic or value-guided, and none considers control feasibility when generating references, which our static feasibility test, descending from static optimization~\cite{crowninshield1981static}, measures.
Synergy- and prior-based methods reduce the action space~\cite{berg2023sar,caggiano2023myodex,he2024dynsyn}, whereas QFlex~\cite{wei2026qflex} explores the full 700-dimensional space, and concurrent MS-Emulator extends the same flow in directions complementary to ours (Sec.~\ref{sec:benchmark}).
NSE keeps the full action space and organizes exploration through the null space of a local excitation-to-force mapping, changing neither the objective nor the action parameterization.

\subsection{Structured and Null-Space Exploration}
Task-space and null-space decomposition is classical in redundant robot control~\cite{khatib1987osc}, and null-space policies have been learned from demonstrations under unknown task constraints~\cite{towell2010nullspace}.
Motor neuroscience shows that biological systems channel variability into task-irrelevant subspaces, regulate it, and exploit it for learning~\cite{scholz1999ucm,todorov2002ofc,davella2003synergies,wu2014variability,dhawale2017variability,latash2012abundance}.
Structured exploration through self-organizing noise~\cite{schumacher2023deprl}, latent-state noise~\cite{chiappa2023lattice}, colored noise~\cite{eberhard2023pink}, state-dependent exploration~\cite{raffin2022gsde}, or parameter-space noise~\cite{plappert2018parameter} shapes the statistics of the noise but not which directions preserve the task.
NSE derives a local null space from a one-control-frame excitation-to-force mapping, applies bounded perturbations to anatomical actuator groups that persist across control steps, and schedules their strength with a two-rate gate.

\section{Problem Formulation}
\label{sec:formulation}
Given a human motion or video sequence $I_{1:T}$, we construct an MS-Human reference 
\begin{equation} \Rset=\{\qref_t,\dot{\bm{q}}^{\star}_t,\bm{c}^{\star}_t,\bm{p}^{\star}_t\}_{t=1}^{T},
\label{eq:reference}
\end{equation}
with reference configuration $\qref_t$, foot-contact schedule $\bm{c}^{\star}_t$, and semantic task sites $\bm{p}^{\star}_t$ (root-relative anatomical joint centers). 
The MS-Human configuration $\bm{q}\in\mathbb{R}^{85}$ contains $42$ polynomial equality couplings, leaving $43$ independent coordinates $\bm{y}\in\mathbb{R}^{43}$ ($6$ root $+37$ internal) with embedding $\bm{q}=\phi(\bm{y})$.
A policy outputs a 700-dimensional action
\begin{equation}
\bm{a}_t\sim\pi_\theta(\,\cdot\mid \bm{s}_t),\qquad \bm{a}_t\in[-1,1]^{700},
\label{eq:action}
\end{equation}
mapped affinely to muscle excitations $\bm{u}_t=\tfrac{1}{2}(\bm{a}_t+\mathbf{1})\in[0,1]^{700}$.
Control runs at $50\,$Hz (ten $2\,$ms simulation substeps per control frame).  

The base learning algorithm is QFlex~\cite{wei2026qflex}, an off-policy actor--critic that transports samples from a learned Gaussian source through a value-guided flow and explores the full $700$-dimensional action space without distinguishing action directions by their short-term mechanical effects.
We keep its observations, rewards, and termination unchanged, so MAR modifies only the reference and NSE only the exploration, isolating their effects from reward redesign.


\section{Muscle-Aware Retargeting}
\label{sec:mar}

Muscle activation dynamics make reference artifacts more costly than in torque-actuated systems, since high-frequency errors require larger excitations that can violate activation bounds and persistent tracking errors accumulate in the discounted return (Appendix~\ref{sec:amplification}).
MAR therefore repairs reference trajectories before policy learning.

\subsection{Faithful Reference Construction}
Four stages map the source motion onto MS-Human-700.

\noindent\emph{1) Semantic joint-center fitting.}
We represent the source motion by $18$ semantic joint centers rather than dense surface markers.
For site $j$ on body $b_j$ with a calibrated local offset $\bm{o}_j\in\mathbb{R}^3$, its position at configuration $\bm{q}$ is
\begin{equation}
\bm{p}_j(\bm{q})=\bm{x}_{b_j}(\bm{q})+\bm{R}_{b_j}(\bm{q})\,\bm{o}_j ,
\label{eq:site}
\end{equation}
where $\bm{x}_{b_j}(\bm{q})\in\mathbb{R}^3$ and $\bm{R}_{b_j}(\bm{q})\in SO(3)$ are the world-frame position and orientation of body $b_j$.
A per-frame trust-region least-squares solver fits these sites to the source joint centers with per-joint weights and temporal regularization.

\noindent\emph{2) Anatomically constrained coordinate reconstruction.}
The optimization operates only on the independent coordinates $\bm{y}$, and the embedding $\bm{q}=\phi(\bm{y})$ reconstructs the dependent coordinates inside every residual evaluation, so every iterate satisfies the polynomial couplings of MS-Human-700.

\noindent\emph{3) Temporal resampling.}
The trajectory is resampled to the $50\,$Hz control clock by shape-preserving interpolation, which avoids filter ringing near contact transitions.

\noindent\emph{4) Contact-aware grounding.}
Grounding aligns the resampled trajectory with a pressure-derived contact schedule through bounded edits of the root height and lower-limb coordinates, with ground contact modeled by a multi-capsule foot and a friction-pyramid basis, and for each frame it solves
\begin{equation}
\begin{aligned}
\bm{y}^{0}_t=\mathop{\arg\min}_{\bm{y}_t\in\mathcal{B}}\;
&\Big\|\tfrac{\bm{p}^{\mathrm{root}}(\bm{y}_t)-\bm{p}^{\star}_t}{\sigma_{\mathrm{task}}}\Big\|^{2}
+\Big\|\tfrac{\bm{\delta}_t}{\sigma_{\mathrm{edit}}}\Big\|^{2}\\
&+\lambda_T\Big\|\tfrac{\bm{\delta}_t-\bm{\delta}_{t-1}}{\sigma_{\mathrm{temp}}}\Big\|^{2}
+\Big\|\tfrac{g(\bm{y}_t)-g^{\star}}{\sigma_g}\Big\|^{2},
\end{aligned}
\label{eq:grounding}
\end{equation}
where $\bm{y}_t$ denotes the independent-coordinate vector at time $t$, $\mathcal{B}$ is its admissible bounded domain, and $\bm{p}^{\mathrm{root}}(\bm{y})$ stacks the root-relative positions of the $18$ sites at $\bm{q}=\phi(\bm{y})$.
The scale factors $\sigma_{\mathrm{task}}$, $\sigma_{\mathrm{edit}}$, $\sigma_{\mathrm{temp}}$, and $\sigma_g$ normalize the residuals and set their relative penalties.

The problem is subject to hard non-penetration, support-band, flight-clearance, and edit-step constraints.
The task scale is much tighter for the upper body than for the lower body, which pins the semantic content of the motion while the legs adapt, the edit is 
$\bm{\delta}_t=\bm{y}_t-\bar{\bm{y}}_t$ with temporal weight $\lambda_T$ and the resampled independent-coordinate reference before contact-aware grounding $\bar{\bm{y}}_t$.
$g$ measures the scheduled support-endpoint gap against a small positive target $g^{\star}$.

\emph{Baseline.}
Our baseline retargeting (BA) follows the conventional recipe for torque-actuated humanoids~\cite{gleicher1998retarget,he2024h2o,fu2024humanplus}, fitting world-frame keypoints with temporal smoothing and using no muscle quantity.
Starting from a Savitzky--Golay-smoothed trajectory, it optimizes the edits $\bm{\delta}_t$ as
\begin{equation}
\begin{aligned}
\min_{\{\bm{\delta}_t\}}\quad
&\sum_t\tfrac12\big\|W^{\frac12}\big(\bm{p}^{\mathrm{w}}(\bm{y}_t)-\bm{p}^{\mathrm{w}\star}_t\big)\big\|^2
+\lambda_1\sum_t
\Big\|\tfrac{\bm{\delta}_t-\bm{\delta}_{t-1}}{\Delta}\Big\|^2 \\
\text{s.t.}\quad
&\|\bm{\delta}_t-\bm{\delta}_{t-1}\|_\infty\le\Delta ,
\end{aligned}
\label{eq:baseline}
\end{equation}
where $\bm{p}^{\mathrm{w}}$ stacks the world-frame site positions of Eq.~\eqref{eq:site}, $\bm{p}^{\mathrm{w}\star}_t$ holds their targets, and $W$ is a diagonal matrix that assigns one weight to each keypoint.
A minimal support rule requires at least one capsule endpoint of each loaded foot inside the support band and none below the floor.

\subsection{Constrained Trajectory Editing}
\label{sec:mar-edits}
Starting from the grounded reference $\bm{y}^{0}_t$ of Eq.~\eqref{eq:grounding}, MAR searches over bounded edits to the independent coordinates
\begin{equation}
\bm{y}_t
=
\bm{y}^{0}_t+\bm{M}\bm{\Phi}_t\bm{Z}.
\label{eq:edits}
\end{equation}

Here, $\bm{\Phi}_t$ is a clamped cubic B-spline basis with few control points, yielding low-dimensional, smooth edits, $\bm{Z}$ contains the knot edits, and $\bm{M}$ restricts editing to lower-limb and shoulder coordinates.
The root remains fixed, and proposals that violate the prescribed per-frame bounds are rejected rather than clipped.
Deterministic coordinate descent over knot, coordinate, and sign continues until coordinate-wise stationarity, accepting an edit only if it improves root feasibility, root-relative task fidelity, or jerk RMS in that priority order.
Thus, fidelity is never traded for smoothness, and MAR minimizes the jerk associated with the high-frequency artifacts analyzed in Eq.~\eqref{eq:amplification}. 

Root feasibility is the summed row-normalized $L_1$ residual of the six root rows of frame-wise inverse dynamics under admissible contact forces. 
Task fidelity is the site-weighted RMS error of root-relative joint centers against the faithful reference, and jerk RMS is the RMS Savitzky--Golay third derivative of the independent coordinates.
Each candidate is screened by a hard-failure vector, preventing improvements in the soft objective from offsetting constraint violations.
The vector checks finiteness, equality residuals, joint ranges, root preservation, velocity, acceleration, and jerk limits, contact and flight constraints, \emph{root-relative} task-fidelity bounds, and a native-contact replay pre-pass.
Only candidates whose vector is identically zero proceed to the soft comparison.

\section{Null-Space Exploration}
\label{sec:nse}


NSE brings a principle of human motor control into policy learning.
Humans express more motor variability along directions that leave task variables unchanged, the \emph{uncontrolled manifold}~\cite{scholz1999ucm}, consistent with \emph{minimal intervention} in optimal feedback control~\cite{todorov2002ofc}. 

This variability is actively regulated, predicts learning performance~\cite{wu2014variability,dhawale2017variability}, and reframes redundancy as \emph{motor abundance}~\cite{latash2012abundance}.
For a 700-muscle policy, NSE estimates excitation directions that minimally affect short-horizon mechanics and focuses exploration on them.



\subsection{Local Excitation-to-Force Mapping}
NSE constructs a local excitation-to-force mapping for the selected anatomical group at each intervention.
For a group containing $n_b$ muscles, $\bm{u}_t\in\mathbb{R}^{n_b}$ and $\bm{\alpha}_{t+1}\in\mathbb{R}^{n_b}$ denote their excitations and endpoint activations, respectively.
The resulting generalized actuator forces are denoted by $\bm{\tau}^{\mathrm{act}}_{t+1}\in\mathbb{R}^{85}$. 
NSE defines the local excitation-to-force mapping $\bm{B}_t\in\mathbb{R}^{85\times n_b}$ as
\begin{equation}
\bm{B}_t
=
\frac{\partial \bm{\tau}^{\mathrm{act}}_{t+1}}
     {\partial \bm{u}_t}
=
\left.
\frac{\partial \bm{\tau}^{\mathrm{act}}}
     {\partial \bm{\alpha}}
\right|_{t+1}
\frac{\partial \bm{\alpha}_{t+1}}
     {\partial \bm{u}_t}.
\label{eq:Bt}
\end{equation}
The activation-transport factor is the exact tangent of the simulator's activation update over ten $2\,$ms substeps and agrees with the simulator to machine precision.
The force factor is the instantaneous activation-to-force mapping at the frame endpoint and combines the moment arms with the force--length--velocity gains.
All $85$ generalized coordinates are retained as rows of $\bm{B}_t$, while its columns are restricted to the selected anatomical group.


\begin{figure}[t]
  \centering
  \includegraphics[width=\linewidth]{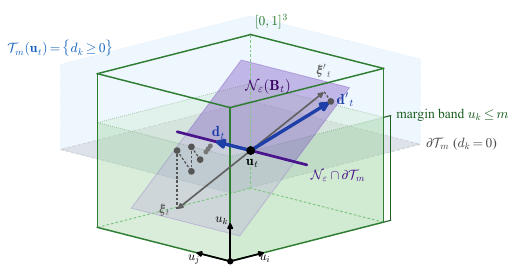}
   \caption{Null directions in the excitation box of three muscles, with an exaggerated margin $m$ (green band).
  With only $u_{t,k}$ within the margin of its lower bound, the cone $\mathcal{T}_m(\bm{u}_t)$ is the half-space $d_k\geq0$ above the gray plane $\partial\mathcal{T}_m$ through $\bm{u}_t$.
  Its intersection with the null space $\mathcal{N}_{\varepsilon}(\bm{B}_t)$ is a half-plane.
  For the sample $\bm{\xi}_t$, Dykstra's iterates (dashed) alternate between $\partial\mathcal{T}_m$ and $\mathcal{N}_{\varepsilon}(\bm{B}_t)$ and converge to the boundary direction $\bm{d}_t$, whereas one projection takes the opposite sample $\bm{\xi}'_t=-\bm{\xi}_t$ to $\bm{d}'_t$ inside the half-plane.}
  \label{fig:dykstra}
\end{figure}

\subsection{Box-Constrained Null-Space Exploration}

NSE seeks excitation perturbations that lie in the numerical null space of the local excitation-to-force mapping while respecting the excitation bounds.
Let
$\bm{B}_t=\bm{U}_t\bm{\Sigma}_t\bm{V}_t^{\top}$
be the singular value decomposition of $\bm{B}_t$.
We form $\bm{V}_{r,t}$ from the right singular vectors whose singular values satisfy $\sigma_i>\varepsilon$.
The numerical null-space projector is then
\begin{equation}
\bm{P}_{\mathcal{N},t}
=
\bm{I}-\bm{V}_{r,t}\bm{V}_{r,t}^{\top},
\qquad
\mathcal{N}_{\varepsilon}(\bm{B}_t)
=
\left\{
\bm{d}\ \middle|\ 
\bm{V}_{r,t}^{\top}\bm{d}=\bm{0}
\right\}.
\label{eq:nullproj}
\end{equation}
Excitations are bounded by
$\bm{\ell}\leq\bm{u}_t\leq\bm{h}$,
where $\bm{\ell}=\bm{0}$ and $\bm{h}=\bm{1}$.
We define a margin-aware feasible-direction cone using the boundary margin $m$ as
\begin{equation}
\mathcal{T}_m(\bm{u}_t)
=
\left\{
\bm{d}\ \middle|\
\begin{array}{ll}
d_i\geq 0 & \text{if } u_{t,i}\leq \ell_i+m,\\
d_i\leq 0 & \text{if } u_{t,i}\geq h_i-m
\end{array}
\right\}.
\label{eq:cone}
\end{equation}
We scale $m$ with the nominal per-muscle perturbation $s/\sqrt{n_b}$.
Near a bound, it permits only inward perturbations and therefore allows saturated muscles to move back toward the interior.


We approximate projection onto
$\mathcal{N}_{\varepsilon}(\bm{B}_t)\cap\mathcal{T}_m(\bm{u}_t)$
using a fixed number of Dykstra-style iterations~\cite{dykstra1983} (Fig.~\ref{fig:dykstra}).
For an initial Gaussian direction $\bm{\xi}_t$, the iterations are
\begin{equation}
\begin{aligned}
\bm{v}^{(k)}
&=
\Pi_{\mathcal{T}_m(\bm{u}_t)}\big(\bm{d}^{(k)}+\bm{e}^{(k)}\big),
&
\bm{d}^{(k+1)}
&=
\bm{P}_{\mathcal{N},t}\bm{v}^{(k)},\\
\bm{e}^{(k+1)}
&=
\bm{d}^{(k)}+\bm{e}^{(k)}-\bm{v}^{(k)},
&
\bm{d}^{(0)}
=
&\bm{\xi}_t,
\qquad
\bm{e}^{(0)}=\bm{0}.
\end{aligned}
\label{eq:dykstra}
\end{equation}
After the final iteration, we project once more onto the numerical null space and remove any remaining outward boundary components.
We denote the resulting approximate projection operator by $\widehat{\Pi}_t$.


Let $s$ denote the target perturbation norm and let $\kappa_t\in(0,1]$ be the largest scalar that keeps the perturbed excitation inside the box.
The executed perturbation is
\begin{equation}
\bm{d}_t
=
\kappa_t s
\frac{\widehat{\Pi}_t(\bm{\xi}_t)}
     {\left\|\widehat{\Pi}_t(\bm{\xi}_t)\right\|}.
\label{eq:null}
\end{equation}
The cone projection is positively homogeneous, so the direction does not depend on the norm of the Gaussian sample, and unlike a direct projection onto the step box $[\bm{\ell}-\bm{u}_t,\bm{h}-\bm{u}_t]$ it avoids the state-dependent clipping of muscles that rest near zero.



\subsection{Anatomical Grouping and Temporal Persistence}
We partition the muscle actuators into ten disjoint anatomical groups based on their moment-arm support, including the left and right legs, upper limbs, lumbopelvic region, thorax, and neck, a coarse mechanically defined analogue of muscle-synergy organization~\cite{davella2003synergies}.
NSE cycles through the groups in round-robin order and intervenes on an expected fraction $p$ of the parallel environments at each step.
A selected environment retains its intervention for three control steps and reprojects the same Gaussian direction through the null space of the current state, which yields temporally coherent variation rather than white noise.


\subsection{Action Perturbation and Replay}

NSE adds the bounded direction $\bm{d}_t$ of Eq.~\eqref{eq:null} to the output of the unmodified value-guided flow immediately before execution, so environments without an intervention behave exactly as in pure QFlex and the perturbation is never reshaped by the flow.
The direction is computed in excitation space and mapped back through the affine map of Sec.~\ref{sec:formulation}, so the action changes by $2\bm{d}_t$.
Every transition is stored with the executed perturbed action, so the critic and flow are trained on the actual behavior distribution without replay filtering.


\subsection{Adaptive Intervention Schedule}
\label{sec:nse-inst}
The base intervention fraction is $p=0.12$ and the excitation-space scale $s=0.012$, shared with the periodic-locomotion study.
NSE modulates its strength with a \emph{two-rate gate} inspired by the fast and slow processes of human motor adaptation~\cite{smith2006tworate}.
The fast state $F_t\in[0,1]$ tracks the pre-exploration uncertainty $\nu_t\in[0,1]$, which combines the RMS spread of flow-action samples with the disagreement between the two $Q$-functions, through its running mean $\bar{\nu}_t$ relative to the initial value $\nu_0$,
\begin{equation}
F_{t+1} = F_t + \alpha_F \left[\operatorname{sigmoid} \left(\frac{\bar{\nu}_t/\nu_0-c}{\tau}\right)-F_t \right],
\label{eq:gatefast}
\end{equation}
where $c$ and $\tau$ set the center and slope of the logistic transition and $\alpha_F$ the update rate, so $F$ starts at one and decays as the policy and critic stabilize.

The slow state $S_t\in[0,1]$ accumulates evidence of competence from the worst recent episodes.
For each metric $X\in\{R,L\}$, namely episode return and length, let $m_{0,X}$ and $s_{0,X}$ be the median and the $10$--$90\%$ spread over the first $N_0$ episodes, $q_{10,X}$ the lower decile over the latest $W$ episodes, and $X^{\uparrow}$ the running upper decile, giving the competence score
\begin{equation}
\pi_X = \operatorname{clip} \left(\frac{q_{10,X}-m_{0,X}}{\max\!\left(X^{\uparrow}-m_{0,X},\,3s_{0,X}\right)}, 0,1 \right).
\label{eq:competence}
\end{equation}
With $C_t=\sqrt{\mu(\pi_R)\mu(\pi_L)}$ and the minimum-jerk profile $\mu(x)=10x^3-15x^4+6x^5$, the slow state follows
\begin{equation}
S_{t+1} = S_t+\alpha_S C_t(1-S_t),
\label{eq:gateslow}
\end{equation}
starting at zero and never decaying.
The two states set the intervention weight
\begin{equation}
w_t = \mu\!\left( 1-(1-F_t)\bigl(1-\mu(S_t)\bigr)\right)\in[0,1],
\label{eq:gate}
\end{equation}
which scales both $p$ and $s$ and yields a high--low--high schedule.
The fast state concentrates exploration while the policy is still uncertain, and the slow state restores it once competence is established, so late training keeps exploring and stays robust.
NSE otherwise leaves the QFlex flow unchanged and adds computation only to intervened environments, and its throughput is about $95\%$ of pure QFlex, so comparisons at equal sample counts are wall-clock-fair to within about $5\%$ of training time.

\section{Experiments}
\label{sec:experiments}


\subsection{Setup}
\label{sec:benchmark}
We ask four questions.
\textbf{RQ1}, does MAR improve muscle-driven feasibility while preserving motion fidelity?
\textbf{RQ2}, does NSE improve sample efficiency, final performance, and training stability over pure QFlex?
\textbf{RQ3}, do MAR and NSE provide complementary benefits?
\textbf{RQ4}, can the complete pipeline train whole-body muscle policies from in-the-wild monocular video, and does MAR improve its trainability?

\emph{Benchmark.}
We evaluate on a deterministic quality-gated subset of HUMAPS-4D~\cite{humaps4d} covering three subjects and eight non-interactive actions (Table~\ref{tab:inventory}).
Each of the ten takes per subject--action pair is screened for annotation quality, EMG integrity, skeletal validity, and joint representability with criteria provided in the released code, and the longest passing take is kept.
This yields $24$ clips totaling $128.55\,$s with durations from $2.20$ to $7.28\,$s, captured at $120\,$Hz and resampled to the $50\,$Hz control rate.
Policy performance never enters selection, and plantar-pressure measurements provide the contact schedules used in Sec.~\ref{sec:mar}.
S01 is the development subject, S02 the validation subject, and S03 a held-out subject.

\emph{Setting.}
For each reference we independently train four policies, $\{\mathrm{BA},\mathrm{MAR}\}\times\{\text{pure QFlex},\mathrm{NSE}\}$, with identical training settings, so the variants differ only in reference construction and exploration.
QFlex is the backbone because it reports strong performance on MS-Human-700 in the native action space~\cite{wei2026qflex,wei2026msemulator}, and concurrent MS-Emulator changes the simulation platform and reward aggregation, so it lies outside this factorial.
The main experiments use motion-capture references, and Sec.~\ref{sec:videopolicy} evaluates the video-based GVHMR branch.

\emph{Metrics.}
Reference-level metrics are joint-center errors as RMS, P95, and maximum values and static muscle feasibility (Sec.~\ref{sec:refcomparison}), and policy-level metrics are those of Table~\ref{tab:main}, where the best checkpoint is the highest-return checkpoint among those with the longest episode length.
Macro-averages are taken over the $24$ references, and all comparisons are paired by reference.

\begin{table}[t]
\caption{Benchmark clips with categories and durations (s).}
\label{tab:inventory}
\centering
\begin{tabular}{lllccc}
\toprule
ID & Action & Category & S01 & S02 & S03\\
\midrule
A03 & Single-leg balance & static & 5.20 & 4.53 & 6.57\\
A05 & Squats             & static & 5.28 & 5.35 & 7.13\\
A08 & Walking            & locomotion & 7.04 & 7.28 & 6.23\\
A09 & Jogging            & locomotion & 6.22 & 6.93 & 5.41\\
A13 & Side-stepping      & locomotion & 6.35 & 6.06 & 5.91\\
A14 & Jump in place      & dynamic & 4.21 & 2.43 & 4.05\\
A17 & Side jump          & dynamic & 4.45 & 2.20 & 3.22\\
A28 & Lean to the side   & postural & 5.47 & 5.13 & 5.90\\
\bottomrule
\end{tabular}
\end{table}

\subsection{Isolating NSE through Periodic Locomotion}
\label{sec:walking}
\begin{figure*}[t]
  \centering
  \includegraphics[width=\textwidth]{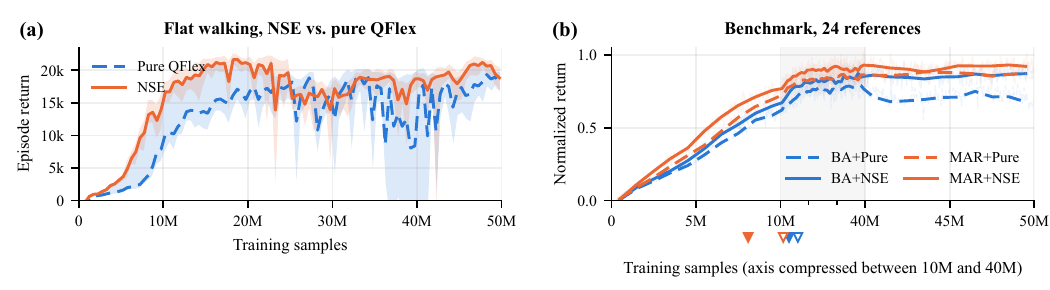}
  \caption{(a) NSE versus pure QFlex on the built-in walking task with the same reference and reward.
  Lines are 500k-bin means of episode return over $50$\,M samples averaged over five seeds, and bands are p10--p90 ranges of the per-episode values pooled across seeds.
  (b) Benchmark learning curves over the first $50$\,M samples, with the axis compressed between $10$ and $40$\,M.
  Episode return is normalized per reference and averaged over the $24$ references, shown as 1M-bin means with faint 50k-step traces.
  Triangles mark the mean first full completion, hollow for pure QFlex and filled for NSE.
  Blue denotes BA and orange MAR, while dashed lines denote pure QFlex and solid lines NSE.}
  \label{fig:walking}\label{fig:curves}
\end{figure*}

Before the factorial study, we isolate NSE on the flat-ground walking task of the QFlex MS-Human-700 environment, where both variants track the same built-in reference under identical rewards and settings and differ only in whether NSE is active.
Fig.~\ref{fig:walking}(a) reports episode return over $50$\,M samples, averaged over five random seeds per variant, as 500k-bin means with p10--p90 bands of the per-episode values pooled across seeds.
NSE reaches a return of $1.5\times10^{4}$ about $6$\,M samples earlier, attains a higher peak return of ${\approx}2.2\times10^{4}$ versus ${\approx}1.9\times10^{4}$, and shows far fewer of the sharp drops that pure QFlex exhibits between $25$ and $45$\,M samples.
Periodicity explains part of this margin, since each episode contains about $5.3$ gait cycles and a recruitment found in one cycle is reused in every later one.

\subsection{Main Results}
\label{sec:mainresults}

Table~\ref{tab:main} and Fig.~\ref{fig:curves}(b) present the factorial comparison, with the sample axis compressed between $10$ and $40$\,M to expose the early gains of MAR and NSE and the late collapse of BA{\,+\,}Pure.
BA{\,+\,}Pure fails to complete the full horizon for three references and BA{\,+\,}NSE for one, and the BA first-completion means count these references at the $50$\,M cap and are therefore lower bounds.

At both exploration levels, MAR reduces the samples to first completion from $16.1$ to $11.0$\,M and from $13.0$ to $8.1$\,M, increases the best return by $1.4$--$2.2$k, and reduces the best joint-center MPJPE by $21$--$39$\,mm (RQ1).
NSE advances first completion by $3.1$\,M samples with BA and $2.9$\,M with MAR, increases the best return by $0.5$--$1.2$k, and reduces the best MPJPE by about $3$\,mm with MAR, and over the trailing window it adds $760$ return and reduces MPJPE by $3.8$\,mm and RMSE by $4.9$\,mm.

No NSE run collapses, whereas BA{\,+\,}Pure collapses late for $6$ of the $24$ references (RQ2).
The effects of MAR and NSE are approximately additive and MAR{\,+\,}NSE performs best in every column (RQ3), while the $3$\,M-sample gain from NSE on finite clips is smaller than the $6$\,M gain on periodic walking, as the periodicity argument of \mbox{Sec.~\ref{sec:walking}} predicts.

\begin{table*}[t]
\caption{$2\times2$ results on the 24-clip benchmark (means over references). 
}
\label{tab:main}
\centering
\footnotesize\setlength{\tabcolsep}{3pt}
\begin{tabularx}{\textwidth}{l*{8}{>{\centering\arraybackslash}X}}
\toprule
Method & Success & Rollout & 1st full & Return & MPJPE & RMSE & qpos MAE & qpos RMSE\\
 & (\%) & success (\%) & (M) & (b/t) & (b/t) & (b/t) & (b/t) & (b/t)\\
\midrule
BA{\,+\,}Pure$^{\ddagger}$ & 87.5 & 76.6 & 16.1 & 18039\,/\,$\dagger$ & 95.4\,/\,$\dagger$ & 129.9\,/\,$\dagger$ & 0.217\,/\,$\dagger$ & 0.294\,/\,$\dagger$\\
BA{\,+\,}NSE$^{\ddagger}$  & 95.8 & 87.8 & 13.0 & 19281\,/\,18243 & 74.2\,/\,78.1 & 97.0\,/\,102.0 & 0.210\,/\,0.219 & 0.281\,/\,0.289\\
MAR{\,+\,}Pure & \textbf{100} & 91.5 & 11.0 & 20231\,/\,19091 & 56.5\,/\,62.3 & 72.6\,/\,80.4 & 0.194\,/\,0.206 & 0.259\,/\,0.275\\
\textbf{MAR{\,+\,}NSE} & \textbf{100} & \textbf{98.5} & \textbf{8.1} & \textbf{20690}\,/\,\textbf{19853} & \textbf{53.7}\,/\,\textbf{58.5} & \textbf{69.3}\,/\,\textbf{75.5} & \textbf{0.190}\,/\,\textbf{0.196} & \textbf{0.252}\,/\,\textbf{0.262}\\
\bottomrule
\multicolumn{9}{@{}p{\textwidth}@{}}{\scriptsize Success is the share of references reaching a full-horizon trajectory, 1st full the samples to the first full-horizon trajectory (M), and b/t the best checkpoint\,/\,the mean over the final $5$\,M samples. Rollout success is the share of $100$ rollouts of the best checkpoint per reference that reach the full horizon with actions sampled from the learned source, $\bm{a}_0=\tanh(\bm{\mu}+\bm{\sigma}\odot\bm{z})$ with $\bm{z}\sim\mathcal{N}(\bm{0},\bm{I})$, and transported through the 20-step flow, with no noise added to the final action. MPJPE and RMSE are joint-center errors (mm), and qpos MAE and RMSE joint-angle errors (rad).
$\dagger$\,collapsed (trailing return below $50\%$ of best) on 6/24 references including the 3 never solved, so trailing values are undefined.\quad
$\ddagger$\,best-checkpoint values include forced-continuation rollouts on the references never solved.}
\end{tabularx}
\end{table*}



\subsection{Ablations of NSE Components}
The factorial comparison ablates MAR and NSE, whereas the components of NSE were fixed by diagnosing why early variants accelerated learning only inconsistently.
Direct clipping to the step box, our first projection, leaks $0.246$ of the force effect, measured as $\|\bm{B}_t\bm{d}_t\|/\|\bm{B}_t\bm{\xi}_t\|$ over four candidate directions at each of three states, so its perturbations were not null in practice, whereas the executed perturbation of the cone projection, after the final null-space projection and boundary cleanup, leaks $2.62\times10^{-9}$ and violates no bound.
Without anatomical grouping, return and episode length at matched steps before the first full trajectory were $17.46\%$ and $16.78\%$ lower.
Without the early decay of the fast state, pure QFlex overtook NSE once it reached $70\%$ of its peak return and length, because continued perturbation then hindered task-space learning, and without the late re-strengthening by the slow state, transient collapses in which a bin's mean return falls below $70\%$ of the peak occurred roughly three times more often.

\begin{figure*}[t]
  \centering
  \includegraphics[width=\textwidth]{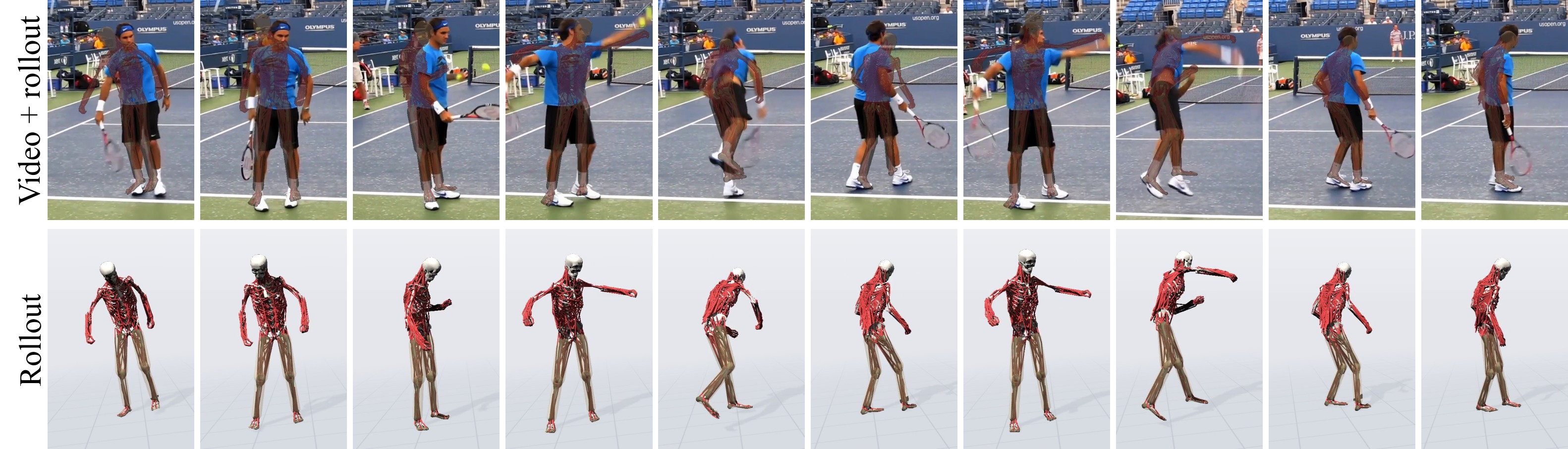}
  \caption{Video to policy on the tennis clip.  Rows show the MAR{\,+\,}NSE rollout overlaid on the video and the rollout alone, sampled about once per second over the $10.38$\,s clip.}
  \label{fig:showcase}
\end{figure*}

\subsection{Reference-Level Comparison}
\label{sec:refcomparison}
Table~\ref{tab:refmetrics} compares the BA and MAR references over the $24$ clips.
Both are equally faithful on average, yet MAR markedly improves \emph{static muscle feasibility}.

The feasibility test asks for each frame whether bounded activations $\bm{\alpha}\in[0,1]^{700}$ and admissible contact forces at the scheduled support feet satisfy the equality-reduced inverse dynamics exactly, treating frames independently without activation dynamics and recording the minimum residual generalized force when they do not.
MAR yields $10.8\times$ more feasible frames than BA ($3.64\%$ versus $0.34\%$), is never worse on any clip, and leads at every residual tolerance (Table~\ref{tab:tolerance}), where $\varepsilon$ bounds the row-normalized $L_1$ residual with root force scaled by body weight, root torque by body weight $\times\,0.2\,$m, and internal rows by actuation-row norms.
Among infeasible frames, MAR lowers the median residual root force from $327$ to $110\,$N, or from $55\%$ to $19\%$ of body weight, and every one of the $1030$ BA support frames lacking a near-ground foot endpoint is repaired by MAR.

\begin{table}[t]
\caption{Reference-level comparison.  Kinematic errors are relative to the faithful reference, static feasibility uses bounded activations with admissible contact forces, and residual is the minimum norm of the unmet generalized force (body weight $590\,$N).}
\label{tab:refmetrics}
\centering

\begin{tabularx}{\columnwidth}{@{}Xcc@{}}
\toprule
Metric & BA & MAR\\
\midrule
Joint-center RMS (mm) & 6.55 & \textbf{6.50}\\
Joint-center P95 (mm) & 26.20 & \textbf{25.94}\\
Joint-center max (mm) & 43.89 & \textbf{40.11}\\
\midrule
Static-feasible frames (\%) & 0.34 & \textbf{3.64}\\
Support frames without near-ground endpoint & 1030 & \textbf{0}\\
Residual root force, median (N) & 327 & \textbf{110}\\
Residual root torque, P95 (N$\cdot$m) & 67.0 & \textbf{7.4}\\
\bottomrule
\end{tabularx}

\end{table}

\begin{table}[t]
\caption{Static-feasible frames (\%) within residual tolerance $\varepsilon$.
}
\label{tab:tolerance}
\centering
\setlength{\tabcolsep}{4.5pt}
\begin{tabular}{lccccccc}
\toprule
$\varepsilon$ & 0 & 0.01 & 0.03 & 0.1 & 0.3 & 1 & 3\\
\midrule
BA  & 0.34 & 0.87 & 1.78 & 10.4 & 23.2 & 59.5 & 90.8\\
MAR & \textbf{3.64} & \textbf{4.38} & \textbf{6.46} & \textbf{22.9}
    & \textbf{48.2} & \textbf{77.3} & \textbf{95.9}\\
\bottomrule
\end{tabular}
\end{table}


\subsection{Video-to-Policy Pipeline}
\label{sec:videopolicy}

For RQ4, we run the complete pipeline on two in-the-wild monocular videos, the tennis clip distributed with GVHMR and a widely circulated online dance clip.
GVHMR reconstructs world-grounded motion, BA and MAR references are built from the same reconstruction, and all four variants are trained on each clip.

On the tennis clip, the best BA{\,+\,}Pure rollout covers only $4.46$\,s of the $10.38$\,s motion and the best BA{\,+\,}NSE rollout $5.64$\,s, and on the $13.97$\,s dance clip they cover $9.46$\,s and $10.38$\,s.
MAR{\,+\,}Pure learns the tennis clip completely but needs $69.5\%$ more samples than MAR{\,+\,}NSE for its first completion, and it covers $13.16$\,s of the dance clip, whereas MAR{\,+\,}NSE learns both clips completely (Fig.~\ref{fig:showcase} shows the tennis clip).

\section{Conclusion and Future Work}
\label{sec:conclusion}

We presented a motion-to-policy pipeline for whole-body musculoskeletal imitation that exploits muscle redundancy twice, in reference construction and in policy exploration.
MAR improves the muscle-driven feasibility of reference trajectories, NSE guides exploration along directions that approximately preserve short-horizon mechanical effects, and with a world-grounded video front end they form the first complete pipeline from monocular video to a whole-body muscle policy.

The combined method completes every benchmark reference with about half the samples of the baseline and $44\%$ lower joint-center error without late collapse.
Future work will extend NSE to task-specific null spaces, which release larger exploration subspaces and approach, in the limit, the bliss of abundance~\cite{latash2012abundance}, and to tendon-driven musculoskeletal hardware~\cite{kawaharazuka2026muscles}.
The code will be released with the paper.


\appendix
\subsection{Artifact Amplification in Muscle Space}
\label{sec:amplification}

\noindent\emph{Value bias.}
For a corrupted reference $\hat p_t=p^{\star}_t+\eta_t$ under a quadratic tracking reward, a policy near the clean motion incurs $\Delta r_t=-\eta_t^{\top}\Lambda\eta_t$ per step, which the critic integrates into $|\Delta Q|\le\sup_t|\Delta r_t|/(1-\gamma)$, a $100\times$ amplification at $\gamma=0.99$.

\noindent\emph{Control cost.}
A position artifact of amplitude $A$ at frequency $f$ demands acceleration $(2\pi f)^2A$.
A torque robot pays the corresponding torque directly, whereas a muscle drives it through the first-order activation dynamics $\tau_a\dot\alpha+\alpha=u$, whose transfer function $H_a(\omega)=1/(1+j\omega\tau_a)$ transmits only $|H_a(\omega)|=(1+(\omega\tau_a)^2)^{-1/2}$ of the commanded amplitude.
Reproducing a torque variation at $\omega=2\pi f$ therefore requires excitation amplified by $1/|H_a|$ and quadratic control cost amplified by
\begin{equation}
\frac{C_{\mathrm{muscle}}}{C_{\mathrm{torque}}}
\;\approx\;\frac{1}{|H_a(\omega)|^{2}}
=1+(2\pi f\,\tau_a)^2 ,
\label{eq:amplification}
\end{equation}
about $15\times$ at $20\,$Hz for $\tau_a=30\,$ms and about $23\times$ for a one-frame discontinuity at the Nyquist frequency of the $50\,$Hz clock, and the implied excitation can leave the feasible $[0,1]$ range.

\noindent\emph{Gradient variance.}
If a stochastic gradient estimate $\bm{g}\in\mathbb{R}^{d}$ of $\nabla f$ carries independent noise of variance $\sigma^{2}$ in each of its $d$ components, then $\mathbb{E}\|\bm{g}-\nabla f\|^{2}=\sum_{i=1}^{d}\mathrm{Var}(g_i)=d\,\sigma^{2}$, so the expected squared deviation from the true gradient grows linearly with the dimension.
With one noise component per action dimension, a $700$-muscle policy suffers roughly $23\times$ the deviation of a torque humanoid with about $30$ actuated degrees of freedom, so an artifact with a comparable tracking penalty imposes an order-of-magnitude control and learning burden that MAR repairs before training.

\bibliographystyle{IEEEtran}
\bibliography{myref} 

\end{document}